\documentclass[journal]{IEEEtran}
\usepackage{amsmath,amsfonts}
\usepackage{algorithmic}
\usepackage{algorithm}
\usepackage{array}
\usepackage[caption=false,font=normalsize,labelfont=sf,textfont=sf]{subfig}
\usepackage{textcomp}
\usepackage{stfloats}
\usepackage{url}
\usepackage{verbatim}
\usepackage{graphicx}
\usepackage{cite}
\usepackage{cleveref}
\usepackage{xcolor}
\usepackage{multirow}
\usepackage{arydshln}
\begin{document}

\title{A Novel Lightweight Transformer with Edge-Aware Fusion for Remote Sensing Image Captioning}

\author{Swadhin Das, Divyansh Mundra, Priyanshu Dayal, Raksha Sharma}



\maketitle

\begin{abstract}
Transformer-based models have achieved strong performance in remote sensing image captioning by capturing long-range dependencies and contextual information. However, their practical deployment is hindered by high computational costs, especially in multi-modal frameworks that employ separate transformer-based encoders and decoders. In addition, existing remote sensing image captioning models primarily focus on high-level semantic extraction while often overlooking fine-grained structural features such as edges, contours, and object boundaries. To address these challenges, a lightweight transformer architecture is proposed by reducing the dimensionality of the encoder layers and employing a distilled version of GPT-2 as the decoder. A knowledge distillation strategy is used to transfer knowledge from a more complex teacher model to improve the performance of the lightweight network. Furthermore, an edge-aware enhancement strategy is incorporated to enhance image representation and object boundary understanding, enabling the model to capture fine-grained spatial details in remote sensing images. Experimental results demonstrate that the proposed approach significantly improves caption quality compared to state-of-the-art methods.
\end{abstract}


\IEEEpeerreviewmaketitle

\section{Introduction}
Remote Sensing Image Captioning (RSIC) is the task of generating descriptive sentences for remote sensing images. This technique has gained popularity because it effectively bridges the gap between visual data and semantic understanding. It supports various real-world applications (such as urban planning, disaster management, land use monitoring, environmental assessment, and agricultural analysis). By automatically interpreting high-resolution remote sensing images, RSIC supports faster and more accurate decision-making in these domains.

The encoder-decoder~\cite{qu2016deep,lu2017exploring} framework is the most widely adopted approach in RSIC. In this setup, an encoder is first used to extract meaningful information from the input image in the form of features. Then using the information decoder generates a meaningful caption of the image. However, developing an effective RSIC model remains challenging due to several factors. These include the limited availability of high-quality training datasets, the presence of domain-specific vocabulary that is often underrepresented, and the inherent complexity of remote sensing images, which may contain diverse objects, varying scales, and ambiguous scenes.

A major issue in existing RSIC models~\cite{hoxha2020new,hoxha2021novel} is that they rarely leverage low-level visual cues that can complement semantic understanding, particularly in scenes with complex layouts or visually similar objects. Limited efforts have been made to explore how enhancing structural detail at the input level influences caption generation quality. However, remote sensing images frequently contain subtle object boundaries and complex spatial arrangements that are difficult to differentiate using only high-level semantic features. Enhancing structural awareness through additional visual cues may help disambiguate objects with similar appearances and improve scene interpretation. In addition, incorporating structural information helps the model better understand spatial relationships, resulting in more accurate and detailed captions. In this work, a fusion-based technique based on edge detection methods combines the original image and an edge-enhanced version of the image, creating a new fused image. The original and edge-enhanced images are then fed into the encoder by modifying its input layer to accept six channels instead of the standard three.

Recently, transformer-based models~\cite{wu2024trtr,das2025good} have gained increasing popularity in RSIC. Although they show prominent results, one major challenge is the computational cost of the model due to a huge number of parameters. Unlike other tasks (such as image classification) where a single transformer is used, here we need two different transformers. This setup increases the complexity of the overall models and limits the practicality of transformer-based RSIC in real-world applications. Due to the complexity of using dual transformer architectures in RSIC, there is a real-world need for a lightweight transformer design to improve practicality. Transferring knowledge to the lightweight offers a promising direction by enabling effective knowledge transfer from complex models to smaller ones, helping retain performance while reducing computational demands. In this work, we have designed a lightweight transformer encoder by reducing the dimensionality of various layers, and we have used knowledge distillation to transfer the knowledge of the actual model to the lightweight model.

The contribution of the work is described as follows.

\begin{itemize}
    \item A lightweight transformer encoder has been designed using knowledge distillation to reduce the overall complexity of the model while maintaining the accuracy of the generated caption.
    \item A novel edge detection-based fusion strategy has been introduced to enhance image representation by combining semantic and structural features, enabling more precise and detailed remote sensing image captioning.
\end{itemize}

The rest of the paper is structured as follows.~\Cref{literature} reviews the related literature.~\Cref{proposed_method} presents the methodology proposed in this work.~\Cref{experiments} describes the experimental setup and reports the results from multiple perspectives (including ablation studies, quantitative comparisons with baseline methods, qualitative evaluations, and visual examples). Finally,~\Cref{conclusion} provides the conclusion.
\section{Literature Review}
\label{literature}
The encoder-decoder~\cite{qu2016deep} framework is widely used in RSIC where a CNN is used to extract the feature of the image and an RNN (such as LSTM) is used to decode the features into a meaningful caption. Das et al.~[2024]~\cite{das2024textgcn} proposed an encoder-decoder framework that combines TextGCN to capture semantic word relationships and a multilayer LSTM decoder, enhanced with a comparison-based beam search for improved RS image captioning. Li et al.~[2024]~\cite{li2024learning} introduced a visual-semantic co-attention and consensus exploitation framework that enhances RSIC by modeling high-level interactions between semantic concepts and visual features. Li et al.~[2025]~\cite{li2025cd4c} proposed a remote sensing image change captioning framework based on change detection that uses binary masks and dual-stream feature fusion to enhance extraction of foreground change features and the accuracy of the caption. Liu et al.~[2025]~\cite{liu2025semantic} introduced a semantic-spatial feature fusion with dynamic graph refinement (SFDR) method that combines multilevel visual features using CLIP and graph attention to improve contextual relevance in RS image captioning. Sree et al.~[2025]~\cite{sree2025residual} proposed a hybrid RSIC approach that combined ResNet50 and Bi-LSTM to improve spatial and spectral feature learning, demonstrating superior performance across multiple evaluation metrics.

With the introduction of the transformer~\cite{vaswani2017attention}, significant advances have been achieved across various domains in natural language processing. In the RSIC field, various methodologies were adopted using a transformer-based architecture with notable improvement. Li et al.~[2024]~\cite{li2024cross} proposed a cross-modal retrieval and semantic refinement method that improves RSIC by integrating refined semantic tokens and visual features through a Transformer Mapper and cross-modal decoder. Meng et al.~[2024]~\cite{meng2024multiscale} proposed a Multiscale Grouping Transformer (MGT) that combines CLIP latent, dilated convolutions, and a global grouping attention mechanism to improve contextual modeling in RSIC. Wu et al.~[2024]~\cite{wu2024trtr} introduced a dual transformer-based RSIC model using a swin transformer for multiscale visual feature extraction and a transformer language model to generate accurate captions. Yang et al.~[2024]~\cite{yang2024bootstrapping} proposed a two-stage vision language pre-training method using a lightweight interactive Fourier transformer to improve image–text alignment and reduce visual feature redundancy in RSIC. Zhao et al.~[2024]~\cite{zhao2024exploring} introduced a Region Attention Transformer (RAT) model that takes advantage of region-grid features and geometry-aware correlations to enhance object-level attention in RSIC tasks. Bazi et al.~[2025]~\cite{bazi2025ragcap} proposed a Retrieval Enhanced Generation (RAGCap) framework that takes advantage of pre-trained vision language models with similarity-based retrieval to generate domain-specific RS captions without fine-tuning. Das et al.~[2025]~\cite{das2025good} comprehensively evaluated twelve CNN architectures within a transformer-based encoder framework to identify optimal encoders to improve RSIC performance. Hu et al.~[2025]~\cite{hu2025rsgpt} constructed RSICap and RSIEval, two high-quality human-annotated datasets to advance the training and evaluation of large vision-language models in remote sensing image captioning. Zhan et al.~[2025]~\cite{zhan2025skyeyegpt} developed SkyEyeGPT, a multimodal large language model trained on the large-scale SkyEye-968k RS instruction-following dataset, achieving strong performance on various RS vision-language tasks without extra encoding modules.
\section{Proposed Method}
\label{proposed_method}
\begin{figure}
    \centering
    \includegraphics[width=\linewidth]{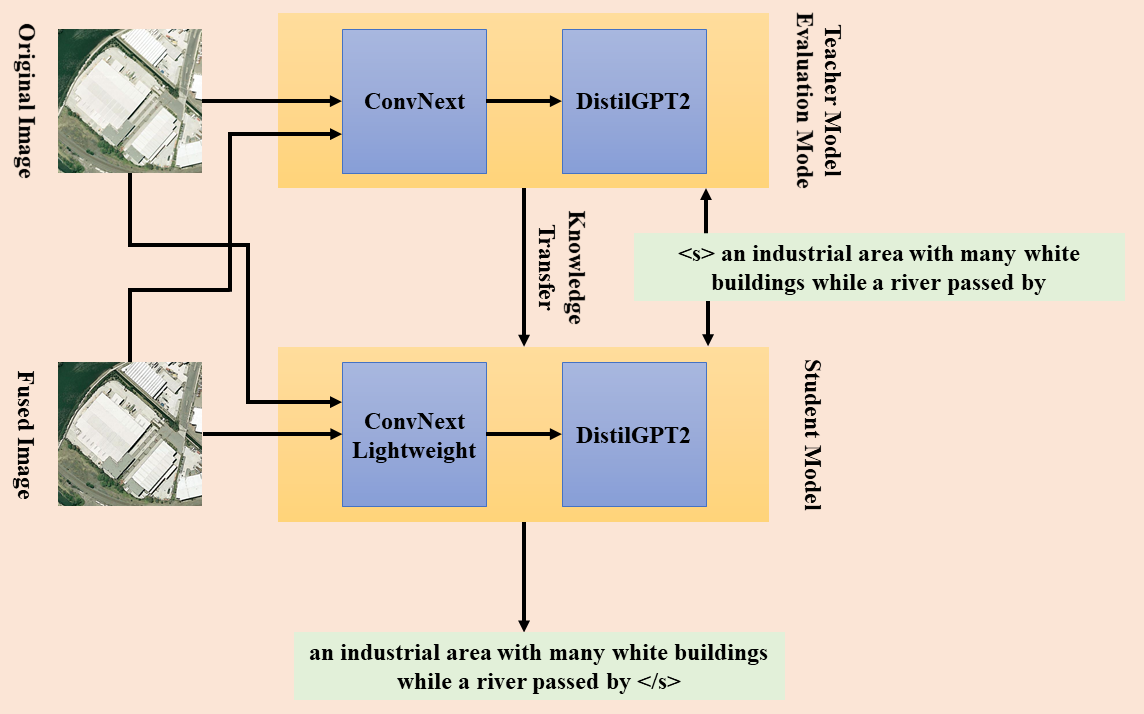}
    \caption{Architecture of the Proposed Model}
    \label{architecture}
\end{figure}

We propose a novel transformer architecture designed to generate meaningful captions from remote sensing images. This work contributes in two key aspects. First, a lightweight model is created by reducing the number of parameters in several layers of the encoder transformer. To enhance its performance, a knowledge distillation approach is employed, transferring knowledge from the full-size teacher model to the lightweight student model. Second, an edge detection-based fusion technique is applied, where edges are accurately detected and fused with the original image to emphasize edge features. As a result, the model takes two input images simultaneously. The detailed architecture of the model is shown in~\Cref{architecture}.
\subsection{Designing of Lightweight Transformer Model}
\begin{table}[!ht]
\centering
\caption{Comparison of bare and distilled models on the SYDNEY dataset}
\label{prf_sydney}
\resizebox{\columnwidth}{!}{
\begin{tabular}{|c|c|c|c|c|c|c|c|c|}
    \hline
    PRF & Model & BLEU-1 & BLEU-2 & BLEU-3 & BLEU-4 & METEOR & ROUGE-L & CIDEr \\
    \hline
    1 & Bare & 0.8094 & 0.6988 & 0.6354 & 0.5808 & 0.4105 & 0.7414 & 2.5345 \\
    \hdashline
    \multirow{2}{*}{2} & Bare & 0.7892 & 0.6899 & 0.6207 & 0.5617 & 0.3873 & 0.7025 & 2.4203 \\
     & Distil & 0.8112 & 0.7183 & 0.6443 & 0.6005 & 0.4272 & 0.7591 & 2.6253 \\
    \hdashline
    \multirow{2}{*}{3} & Bare & 0.7861 & 0.6919 & 0.6261 & 0.5699 & 0.3810 & 0.7099 & 2.5193 \\
     & Distil & \textbf{0.8214} & \textbf{0.7212} & \textbf{0.6515} & \textbf{0.6247} & \textbf{0.4324} & \textbf{0.7749} & \textbf{2.6887} \\
    \hdashline
    \multirow{2}{*}{4} & Bare & 0.7421 & 0.6498 & 0.5850 & 0.5150 & 0.3639 & 0.6822 & 2.2025 \\
     & Distil & 0.7534 & 0.6668 & 0.5906 & 0.5231 & 0.3798 & 0.6983 & 2.2667 \\
    \hdashline
    \multirow{2}{*}{5} & Bare & 0.7240 & 0.6302 & 0.5684 & 0.5015 & 0.3584 & 0.6677 & 2.0876 \\
     & Distil & 0.7326 & 0.6565 & 0.5871 & 0.5170 & 0.3665 & 0.6810 & 2.1465 \\
    \hline
\end{tabular}}
\end{table}
\begin{table}[!ht]
\centering
\caption{Comparison of bare and distilled models on the UCM dataset}
\label{prf_ucm}
\resizebox{\columnwidth}{!}{
\begin{tabular}{|c|c|c|c|c|c|c|c|c|}
    \hline
    PRF & Model & BLEU-1 & BLEU-2 & BLEU-3 & BLEU-4 & METEOR & ROUGE-L & CIDEr \\
    \hline
    1 & bare & 0.8504 & 0.7890 & 0.7416 & 0.6998 & 0.4863 & 0.8006 & 3.4452 \\
    \hdashline
    \multirow{2}{*}{2} & bare & 0.8582 & 0.7976 & 0.7489 & 0.7072 & 0.4902 & 0.8212 & 3.3920 \\
     & distil & \textbf{0.8592} & \textbf{0.8032} & \textbf{0.7611} & \textbf{0.7244} & 0.5057 & \textbf{0.8313} & \textbf{3.5083} \\
    \hdashline
    \multirow{2}{*}{3} & bare & 0.8338 & 0.7651 & 0.7150 & 0.6715 & 0.4616 & 0.8144 & 3.2959 \\
     & distil & 0.8456 & 0.7933 & 0.7519 & 0.7199 & \textbf{0.5089} & 0.8249 & 3.4092 \\
    \hdashline
    \multirow{2}{*}{4} & bare & 0.7858 & 0.7181 & 0.6691 & 0.6294 & 0.4357 & 0.7502 & 2.8403 \\
     & distil & 0.8189 & 0.7477 & 0.6885 & 0.6362 & 0.4325 & 0.7637 & 2.9456 \\
    \hdashline
    \multirow{2}{*}{5} & bare & 0.7352 & 0.6523 & 0.5962 & 0.5523 & 0.3692 & 0.6681 & 2.3986 \\
     & distil & 0.7752 & 0.6991 & 0.6451 & 0.6017 & 0.4195 & 0.7297 & 2.6741 \\
    \hline
\end{tabular}}
\end{table}
\begin{table}[!ht]
\centering
\caption{Comparison of bare and distilled models on the RSICD dataset}
\label{prf_rsicd}
\resizebox{\columnwidth}{!}{
\begin{tabular}{|c|c|c|c|c|c|c|c|c|}
    \hline
    PRF & Model & BLEU-1 & BLEU-2 & BLEU-3 & BLEU-4 & METEOR & ROUGE-L & CIDEr \\
    \hline
    1 & bare & 0.6387 & 0.4594 & 0.3592 & 0.2871 & 0.2587 & \textbf{0.4955} & \textbf{0.8650} \\
    \hdashline
    \multirow{2}{*}{2} & bare & 0.6290 & 0.4518 & 0.3545 & 0.2819 & 0.2501 & 0.4864 & 0.8396 \\
     & distil & \textbf{0.6480} & \textbf{0.4657} & \textbf{0.3638} & \textbf{0.2962} & \textbf{0.2619} & 0.4916 & 0.8579 \\
    \hdashline
    \multirow{2}{*}{3} & bare & 0.6138 & 0.4351 & 0.3286 & 0.2583 & 0.2448 & 0.4572 & 0.7723 \\
     & distil & 0.6262 & 0.4505 & 0.3447 & 0.2725 & 0.2458 & 0.4623 & 0.8137 \\
    \hdashline
    \multirow{2}{*}{4} & bare & 0.6034 & 0.4261 & 0.3220 & 0.2533 & 0.2374 & 0.4435 & 0.7557 \\
     & distil & 0.6208 & 0.4466 & 0.3418 & 0.2701 & 0.2460 & 0.4607 & 0.8013 \\
    \hdashline
    \multirow{2}{*}{5} & bare & 0.5980 & 0.4231 & 0.3196 & 0.2516 & 0.2365 & 0.4463 & 0.7344 \\
     & distil & 0.6160 & 0.4426 & 0.3385 & 0.2685 & 0.2529 & 0.4621 & 0.7971 \\
    \hline
\end{tabular}}
\end{table}
This work develops a lightweight version of the encoder transformer by modifying the convolutional neural network (ConvNext) component within the encoder, which is achieved by applying a Parameter Reduction Factor (PRF). PRF systematically reduces the complexity of the associated components in the CNN. This reduction aims to create a more efficient model with fewer parameters and lower computational requirements, making it suitable for deployment in resource-constrained environments without significantly compromising performance. The reduction approach applied to the convolutional neural network is summarized as follows.
\begin{itemize}
    \item The number of input and output channels in convolutional layers is proportionally decreased according to the PRF, with necessary adjustments made to maintain valid group convolution configurations, except in cases where the input dimension is predefined and fixed.
    \item The number of feature channels in the normalization layers associated with convolutional output is reduced on the basis of the PRF.
    \item Feature-wise normalization layers are modified by scaling down their normalized dimensions according to the PRF.
    \item The output dimensions of fully connected layers are reduced in proportion to the PRF, except in cases where the output dimension is predefined and fixed.
    \item Additional layer-specific parameters, such as internal scaling vectors, are resized in line with the PRF to maintain structural consistency.
    \item These structural modifications are recursively applied to all eligible components within the CNN to achieve uniform parameter reduction while preserving the network's architectural design.
\end{itemize}

To accommodate the changes introduced by these structural modifications, the associated weights and biases in each layer are resized using an interpolation-based strategy. Specifically, multi-dimensional interpolation is employed to scale the original parameters to their new shapes while preserving their relative distribution. This adjustment ensures that the reduced model maintains continuity with the original initialization, facilitating stable training and convergence despite the reduction in capacity.

\Cref{prf_sydney,prf_ucm,prf_rsicd} presents the comparison between the original model and its lightweight counterparts for various values of the Parameter Reduction Factor (PRF), up to $PRF = 5$. In these tables, $PRF=1$ corresponds to the original model without any parameter reduction. The results demonstrate that lightweight models achieve performance comparable to that of the original model despite having significantly fewer parameters. 

In particular, on datasets with limited training data, such as SYDNEY and UCM, lightweight models with up to $PRF=3$ outperform the original model, after which performance begins to decline. In the case of the RSICD dataset, the best performance is observed at $PRF=2$, beyond which the accuracy of the model deteriorates. The effect of the total parameters in the encode with varing PRF is shown below.
\begin{figure}[!ht]
    \centering
    \includegraphics[width=0.45\linewidth]{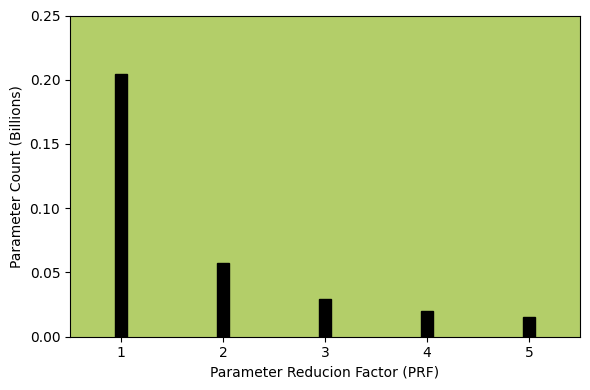}
    \caption{PRF vs Total Parameters}
    \label{fig:PRF}
\end{figure}
The relationship between the Parameter Reduction Factor (PRF) and the total number of parameters in the ConvNext-based transformer encoder is shown in~\Cref{fig:PRF}. As the PRF increases, the models become progressively lighter, yet maintain competitive performance up to a certain threshold. The application of knowledge distillation further enhances the effectiveness of these lightweight models within this range.
\subsection{Edge Detection-based Fusion Technique}
\begin{table}[!ht]
\centering
\caption{Comparison of Different Edge Detected Fusion Models on the SYDNEY dataset}
\label{edge_sydney}
\resizebox{\columnwidth}{!}{
\begin{tabular}{|c|c|c|c|c|c|c|c|}
    \hline
    Edge & BLEU-1 & BLEU-2 & BLEU-3 & BLEU-4 & METEOR & ROUGE-L & CIDEr\\
    \hline
    Original & 0.8214 & 0.7212 & 0.6515 & 0.6247 & 0.4324 & 0.7749 & 2.6887 \\
    Canny~\cite{canny1986computational} & \textbf{0.8358} & \textbf{0.7505} & \textbf{0.6764} & \textbf{0.6498} & \textbf{0.4579} & \textbf{0.7740} & \textbf{2.8633} \\
    Sobel~\cite{sobel19683x3} & 0.8316 & 0.7448 & 0.6710 & 0.6417 & 0.4457 & 0.7630 & 2.8198 \\
    Laplacian~\cite{marr1980theory} & 0.8338 & 0.7490 & 0.6702 & 0.6412 & 0.4471 & 0.7698 & 2.7756 \\
    \hline
\end{tabular}
}
\end{table}
\begin{table}[!ht]
\centering
\caption{Comparison of Different Edge Detected Fusion Models on the UCM dataset}
\label{edge_ucm}
\resizebox{\columnwidth}{!}{
\begin{tabular}{|c|c|c|c|c|c|c|c|}
    \hline
    Edge & BLEU-1 & BLEU-2 & BLEU-3 & BLEU-4 & METEOR & ROUGE-L & CIDEr\\
    \hline
    Original & 0.8592 & 0.8032 & 0.7611 & 0.7244 & 0.5057 & 0.8313 & 3.5083 \\
    Canny~\cite{canny1986computational} & 0.8758 & 0.8098 & 0.7566 & 0.7098 & 0.4949 & 0.8408 & 3.6435 \\
    Sobel~\cite{sobel19683x3} & 0.8775 & 0.8126 & 0.7613 & 0.7179 & 0.5103 & 0.8436 & 3.6241 \\
    Laplacian~\cite{marr1980theory} & \textbf{0.8810} & \textbf{0.8216} & \textbf{0.7675} & \textbf{0.7231} & \textbf{0.5219} & \textbf{0.8502} & \textbf{3.6348} \\
    \hline
\end{tabular}
}
\end{table}
\begin{table}[!ht]
\centering
\caption{Comparison of Different Edge Detected Fusion Models on the RSICD dataset}
\label{edge_rsicd}
\resizebox{\columnwidth}{!}{
\begin{tabular}{|c|c|c|c|c|c|c|c|}
    \hline
    Edge & BLEU-1 & BLEU-2 & BLEU-3 & BLEU-4 & METEOR & ROUGE-L & CIDEr\\
    \hline
    Original & 0.6480 & 0.4657 & 0.3638 & 0.2962 & 0.2619 & 0.4916 & 0.8579 \\
    Canny~\cite{canny1986computational} & 0.6655 & 0.4945 & 0.3907 & 0.3204 & 0.2931 & 0.5249 & 0.8869 \\
    Sobel~\cite{sobel19683x3} & 0.6733 & 0.4966 & 0.3856 & 0.3105 & 0.2896 & 0.5187 & 0.8814 \\
    Laplacian~\cite{marr1980theory} & \textbf{0.6781} & \textbf{0.5019} & \textbf{0.3971} & \textbf{0.3242} & \textbf{0.2984} & \textbf{0.5295} & \textbf{0.9196} \\
    \hline
\end{tabular}
}
\end{table}
This work employs three widely used edge detection techniques: Canny, Sobel, and Laplacian. The technical details of these methods are provided below.
\begin{enumerate}
    \item\textbf{Canny~\cite{canny1986computational}:} The Canny edge detector is a multistage algorithm developed to identify a wide range of edges with precision. It achieves optimal edge detection by balancing noise reduction, localization accuracy, and minimal spurious response.
    \item\textbf{Sobel~\cite{sobel19683x3}:} The Sobel operator is a simple gradient-based method that uses convolutional kernels to estimate changes in intensity in an image. It is widely used for detecting horizontal and vertical edges because of its computational efficiency and effectiveness.
    \item\textbf{Laplacian~\cite{marr1980theory}:} Laplacian edge detection identifies edges by detecting zero-crossings in the second derivative of the image intensity. It is particularly effective in highlighting regions of rapid intensity change after initial smoothing with a Gaussian filter.
\end{enumerate}
\begin{figure}[!ht]
    \centering
    \includegraphics[width=0.45\linewidth]{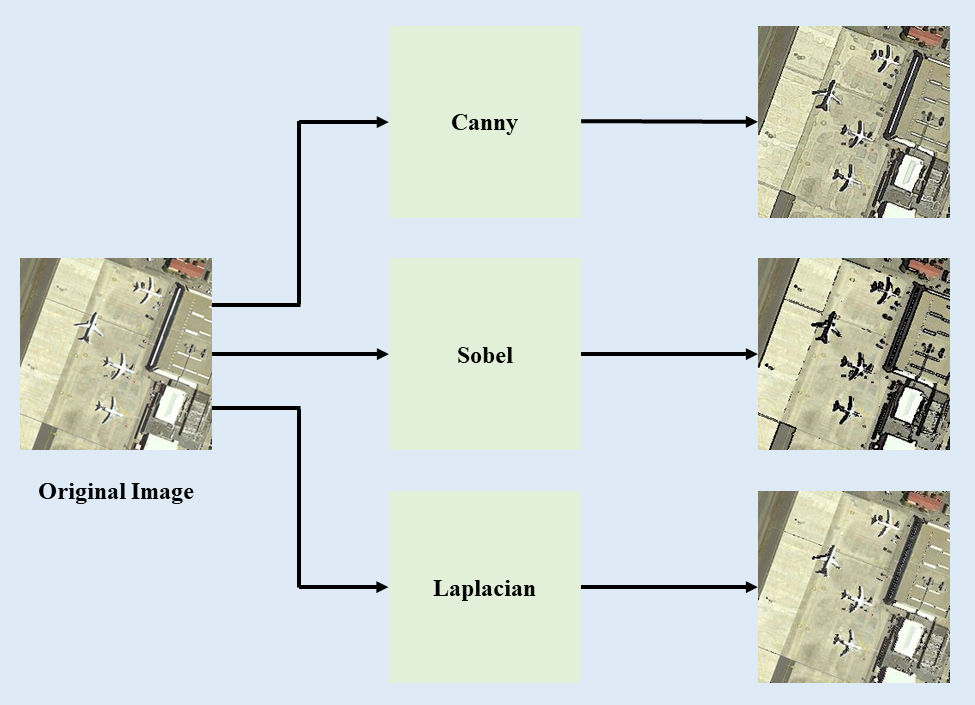}
    \caption{Results of Different Edge Detection Techniques}
    \label{edge_detection}
\end{figure}
The output of different edge detection techniques is shown in~\Cref{edge_detection}.

To effectively leverage the structural information captured by edge detectors, edge maps are extracted from the input images. These edge maps are then fused with the original images, allowing the model to benefit from both intensity-based and boundary-focused representations. Instead of processing original and edge-enhanced images through separate streams, which would significantly increase the model's complexity and contradict the objective of developing a lightweight architecture, we adopt a more efficient approach. Specifically, only the input convolutional layer is modified to accept six channels instead of three, thereby accommodating both the original and edge-enhanced images within a single unified processing pipeline.

~\Cref{edge_sydney,edge_ucm,edge_rsicd} illustrates the impact of incorporating edge-detection-based fused images in our framework. Across all three datasets, a notable improvement in performance is observed when the fused images are used. However, the differences in performance among the individual edge detection methods are relatively small. Still, it can be observed that for the SYDNEY dataset, the Canny edge detector yields slightly better results, whereas for the other two datasets, the Laplacian method performs marginally better than the alternatives.
\section{Experiments}
\label{experiments}
The proposed model adopts a standard transformer-based architecture (see~\Cref{architecture}), in which a lightweight ConvNeXt combined with a multi-head transformer serves as the encoder, and DistilGPT2 is employed as the decoder. All experiments were conducted on a system running Ubuntu $20.04$, equipped with an~\emph{NVIDIA RTX A6000 GPU (50 GB)}. The implementation was carried out using Jupyter Notebook within a Docker container environment. During the preparation of this document, Grammarly and ChatGPT were utilized solely for language refinement, including rephrasing and correction of grammatical and syntactical issues, without introducing any new technical content.
\subsection{Experimental Setup}
The encoder consists of $16$ attention heads and six stacked multi-head transformer layers, with a patch size of $49$. The decoder is equipped with a cross-attention layer, enabling it to incorporate encoder representations effectively. We fine-tuned the decoder using the training dataset before integrating it into our model. Consequently, the size of the vocabulary and the positional embeddings were adjusted accordingly. We compared the performance of our RSIC model using DistilGPT2 against two other decoder variants, with detailed results provided in the supplementary file. For DistilGPT2, we employed the standard configuration of $12$ attention heads and six stacked multi-head transformer layers. The hidden size of the encoder and the decoder is set to $768$ individually, the standard dimension used in transformer architectures.
\subsection{Dataset Used}
In our work, we have used three popular RSIC datasets: SYDNEY~\cite{qu2016deep}, UCM~\cite{qu2016deep}, and RSICD~\cite{lu2017exploring}. Details of these datasets are provided below. 
\begin{enumerate}
    \item\textbf{SYDNEY}~\cite{qu2016deep}: This dataset contains $613$ images, including $497$ for training, $58$ for testing, and $58$ for validation. It is derived from the Sydney dataset~\cite{zhang2014saliency}, a large-scale image collection of the Sydney region in Australia. The dataset was obtained from Google Earth and features images with a resolution of $18{,}000 \times 14{,}000$ pixels and a spatial resolution of 0.5 meters per pixel. It comprises seven classes after cropping and selection:~\emph{airport},~\emph{industrial},~\emph{meadow},~\emph{ocean},~\emph{residential},~\emph{river}, and~\emph{runway}.
    \item\textbf{UCM}~\cite{qu2016deep}: This dataset includes $2100$ images, with $1680$ for training, $210$ for testing, and $210$ for validation. It is based on the UC Merced Land Use dataset~\cite{yang2010bag}, which consists of images with a resolution of 256×256 pixels and a spatial resolution of 0.3048 meters per pixel. These images were manually cropped from larger aerial scenes acquired through the United States Geological Survey (USGS) National Map Urban Area Imagery. It consists of $21$ land use classes, each containing $100$ images. The classes include~\emph{agriculture},~\emph{airport},~\emph{baseball diamond},~\emph{beach},~\emph{buildings},~\emph{chaparral},~\emph{dense-residential},~\emph{forest},~\emph{freeway},~\emph{golf-course},~\emph{harbour},~\emph{intersection},~\emph{medium-residential},~\emph{mobile-home-park},~\emph{overpass},~\emph{parking},~\emph{river},~\emph{runway},~\emph{sparse-residential},~\emph{storage-tanks}, and~\emph{tennis-court}.
    \item\textbf{RSICD}~\cite{lu2017exploring}: This dataset contains $10,\!921$ images, including $8034$ for training, $1093$ for testing, and $1094$ for validation. It is compiled from sources such as Google Earth~\cite{xia2017aid}, Baidu Map, MapABC, and Tianditu. The dataset comprises $31$ classes, including~\emph{airport},~\emph{bareland},~\emph{baseball-field},~\emph{beach},~\emph{bridge},~\emph{center},~\emph{church},~\emph{commercial},~\emph{dense-residential},~\emph{desert},~\emph{farmland},~\emph{forest},~\emph{industrial},~\emph{meadow},~\emph{medium-residential},~\emph{mountain},~\emph{park},~\emph{parking},~\emph{playground},~\emph{pond},~\emph{port},~\emph{railway-station},~\emph{resort},~\emph{river},~\emph{school},~\emph{sparse-residential},~\emph{square},~\emph{stadium},~\emph{storage-tanks}, and~\emph{viaduct}.
\end{enumerate}
However, we have used the modified version of these datasets~\cite{das2024textgcn}, where the authors have corrected many spelling and grammatical errors and solved many issues, such as consistencies between dialects (fixed American English).
\subsection{Peformance Metrices Used}
The proposed method is comprehensively evaluated using seven widely adopted metrics in remote sensing image captioning (RSIC): BLEU-1 to BLEU-4, METEOR, ROUGE-L, and CIDEr. All evaluations are conducted using the~\emph{nlgeval}~\cite{sharma2017nlgeval} package. Descriptions of these metrics are provided below.
\begin{enumerate}
    \item \textbf{BLEU:} BLEU (Bilingual Evaluation Understudy)~\cite{papineni-etal-2002-bleu} is a precision-based metric that evaluates $n$-gram overlap between candidate and reference captions. It includes a brevity penalty to discourage short output. We use BLEU-1 to BLEU-4.
    
    \item \textbf{METEOR:} METEOR (Metric for Evaluation of Translation with Explicit Ordering)~\cite{lavie-agarwal-2007-meteor} is a recall-weighted metric that combines unigram precision and recall with stemming and synonym matching. It is more semantically aware than BLEU but computationally heavier.
    
    \item \textbf{ROUGE:} ROUGE (Recall-Oriented Understudy for Gisting Evaluation)~\cite{lin-2004-ROUGE} is a recall-focused metric that measures $n$-gram and sequence overlap. We use the ROUGE-L variant, which is considered the longest common subsequence.
    
    \item \textbf{CIDEr:} CIDE (Consensus-based Image Description Evaluation)~\cite{vedantam2015cider}, which is specially designed for image captioning, computes the TF-IDF-weighted $n$ gram similarity across multiple references, emphasizing rare but meaningful words. We use CIDEr to assess semantic fidelity.
\end{enumerate}
\section{Results and Analysis}
\label{experimental_results}
The results of the proposed method are presented from multiple perspectives, including numerical comparisons with various state-of-the-art approaches (see~\Cref{comp_rsicd,comp_sydney,comp_ucm}), ablation studies of our work, and visual examples against several transformer architectures. In addition, we have shown the effect of different transformer-based decoders with ConvNext as encoder transformers in RSIC.
\subsection{Effect of Different Decoders in Transformer-based RSIC}
\begin{table}[!ht]
\centering
\caption{Comparison of Different Decoders on the SYDNEY Dataset}
\label{DECODER_SYDNEY}
\resizebox{\columnwidth}{!}{
\begin{tabular}{|c|c|c|c|c|c|c|c|c|}
    \hline
    Decoder & BLEU-1 & BLEU-2 & BLEU-3 & BLEU-4 & METEOR & ROUGE-L & CIDEr & \#PARAMS \\
    \hline
    RoBERTa & 0.7909 & 0.6763 & 0.6298 & 0.5591 & 0.4036 & 0.7193 & 2.5133 & 0.1144 \\
    GPT2~\cite{das2025good} & 0.7997 & 0.6844 & 0.6325 & 0.5694 & 0.4073 & 0.7349 & 2.4945 & 0.1138 \\
    DistilGPT2 & \textbf{0.8094} & \textbf{0.6988} & \textbf{0.6354} & \textbf{0.5808} & \textbf{0.4105} & \textbf{0.7414} & \textbf{2.5345} & \textbf{0.0571}\\
    \hline
\end{tabular}}
\end{table}
\begin{table}[!ht]
\centering
\caption{Comparison of Different Decoders on the UCM Dataset}
\label{DECODER_UCM}
\resizebox{\columnwidth}{!}{
\begin{tabular}{|c|c|c|c|c|c|c|c|c|}
    \hline
    Decoder & BLEU-1 & BLEU-2 & BLEU-3 & BLEU-4 & METEOR & ROUGE-L & CIDEr & \#PARAMS \\
    \hline
    RoBERTa & 0.8409 & \textbf{0.8061} & \textbf{0.7443} & \textbf{0.7087} & 0.4615 & 0.8058 & 3.2185 & 0.1146 \\
    GPT2~\cite{das2025good} & 0.8369 & 0.7712 & 0.7143 & 0.6612 & 0.4566 & \textbf{0.8119} & \textbf{3.4582} & 0.1140 \\
    DistilGPT2 & \textbf{0.8504} & 0.7890 & 0.7416 & 0.6998 & \textbf{0.4863} & 0.8006 & 3.4452 &~\emph{0.0573} \\
    \hline
\end{tabular}}
\end{table}
\begin{table}[!ht]
\centering
\caption{Comparison of Different Decoders on the RSICD Datasets}
\label{DECODER_RSICD}
\resizebox{\columnwidth}{!}{
\begin{tabular}{|c|c|c|c|c|c|c|c|c|}
    \hline
    Decoder & BLEU-1 & BLEU-2 & BLEU-3 & BLEU-4 & METEOR & ROUGE-L & CIDEr & \#PARAMS \\
    \hline
    RoBERTa & 0.6396 & 0.4690 & 0.3674 & 0.2845 & 0.2446 & 0.4829 & 0.8333 & 0.1176 \\
    GPT2~\cite{das2025good} & \textbf{0.6431} & \textbf{0.4665} & \textbf{0.3602} & \textbf{0.3013} & 0.2560 & 0.4945 & 0.8415 & 0.1170 \\
    DistilGPT2 & 0.6387 & 0.4594 & 0.3592 & 0.2871 & \textbf{0.2587} & \textbf{0.4955} & \textbf{0.8650} &~\emph{0.0608} \\
    \hline
\end{tabular}}
\end{table}
The transformer-based RSIC model was assessed using three commonly adopted decoders: RoBERTa, GPT2, and DistilGPT2, each paired with a ConvNext CNN-based encoder.~\Cref{DECODER_RSICD,DECODER_SYDNEY,DECODER_UCM} presents a quantitative comparison of these decoders. Here, the number of parameters in the corresponding decoder (denoted~\emph{\#PARAMS}) is expressed in billions and varies slightly across datasets due to independent fine-tuning. Among the three, DistilGPT2 consistently demonstrated the lowest parameter count while maintaining competitive performance, establishing it as the most efficient and effective decoder in this study.
\subsection{Ablation Studies}
\begin{table}[!ht]
\centering
\caption{Ablation Studies of Our Work on the SYDNEY dataset}
\label{abl_sydney}
\resizebox{\columnwidth}{!}{
\begin{tabular}{|c|c|c|c|c|c|c|c|c|c|}
    \hline
    Edge & BLEU-1 & BLEU-2 & BLEU-3 & BLEU-4 & METEOR & ROUGE-L & CIDEr\\
    \hline
    Conv+DGPT2 & 0.8094 & 0.6988 & 0.6354 & 0.5808 & 0.4105 & 0.7414 & 2.5345 \\
    DConv+DGPT2 & 0.8214 & 0.7212 & 0.6515 & 0.6247 & 0.4324 & 0.7749 & 2.6887 \\
    Conv+DGPT2+Edge & \textbf{0.8417} & \textbf{0.7580} & 0.6654 & 0.6389 & \textbf{0.4613} & \textbf{0.7812} & 2.8132 \\
    DConv+DGPT2+Edge & 0.8358 & 0.7505 & \textbf{0.6764} & \textbf{0.6498} & 0.4579 & 0.7740 & \textbf{2.8633} \\
    \hline
\end{tabular}}
\end{table}
\begin{table}[!ht]
\centering
\caption{Ablation Studies of Our Work on the UCM dataset}
\label{abl_ucm}
\resizebox{\columnwidth}{!}{
\begin{tabular}{|c|c|c|c|c|c|c|c|c|c|}
    \hline
    Edge & BLEU-1 & BLEU-2 & BLEU-3 & BLEU-4 & METEOR & ROUGE-L & CIDEr\\
    \hline
    Conv+DGPT2 & 0.8504 & 0.7890 & 0.7416 & 0.6998 & 0.4863 & 0.8006 & 3.4452 \\
    DConv+DGPT2 & 0.8592 & 0.8032 & 0.7611 & 0.7244 & 0.5057 & 0.8313 & 3.5083 \\
    Conv+DGPT2+Edge & 0.8714 & 0.8132 & \textbf{0.7713} & \textbf{0.7319} & 0.5105 & 0.8392 & 3.5412 \\ 
    DConv+DGPT2+Edge & \textbf{0.8810} & \textbf{0.8216} & 0.7675 & 0.7231 & \textbf{0.5219} & \textbf{0.8502} & \textbf{3.6348} \\
    \hline
\end{tabular}}
\end{table}
\begin{table}[!ht]
\centering
\caption{Ablation Studies of Our Work on the RSICD dataset}
\label{abl_rsicd}
\resizebox{\columnwidth}{!}{
\begin{tabular}{|c|c|c|c|c|c|c|c|c|c|}
    \hline
    Edge & BLEU-1 & BLEU-2 & BLEU-3 & BLEU-4 & METEOR & ROUGE-L & CIDEr\\
    \hline
    Conv+DGPT2 & 0.6387 & 0.4594 & 0.3592 & 0.2871 & 0.2587 & 0.4955 & 0.8650 \\
    DConv+DGPT2 & 0.6480 & 0.4657 & 0.3638 & 0.2962 & 0.2619 & 0.4916 & 0.8579 \\
    Conv+DGPT2+Edge & \textbf{0.6812} & \textbf{0.5062} & \textbf{0.4011} & 0.3209 & 0.2844 & 0.5216 & 0.9070 \\
    DConv+DGPT2+Edge & 0.6781 & 0.5019 & 0.3971 & \textbf{0.3242} & \textbf{0.2984} & \textbf{0.5295} & \textbf{0.9196} \\
    \hline
\end{tabular}}
\end{table}
We perform an ablation study to evaluate the efficiency of different components of our work.~\Cref{abl_sydney,abl_ucm,abl_rsicd} demonstrate the effectiveness of individual components of our work. Here, Conv denotes ConvNext, D in the suffix denotes the distilled version of the corresponding architecture, and Edge denotes that the corresponding experiment is performed with both the input image and the edge-detected version. The results reveal that the lightweight transformer encoder maintains a performance quality that is equivalent to that of the original. In addition, it is noted that adding the edge-detected version significantly improves performance.
\begin{figure*}[!th]
  \centering
  \subfloat[\label{example1}]{\includegraphics[width=0.275\textwidth,height=150px]{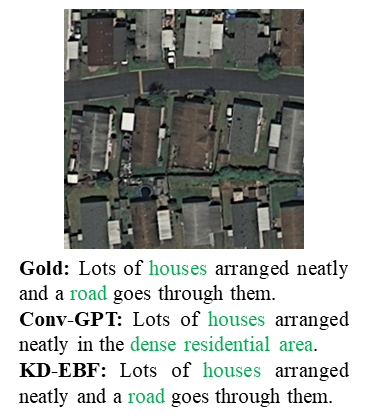}}
  \subfloat[\label{example2}]{\includegraphics[width=0.275\textwidth,height=150px]{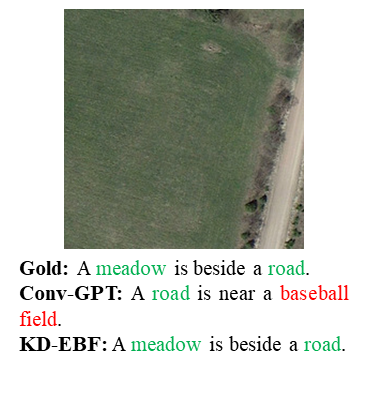}}
  \subfloat[\label{example3}]{\includegraphics[width=0.275\textwidth,height=150px]{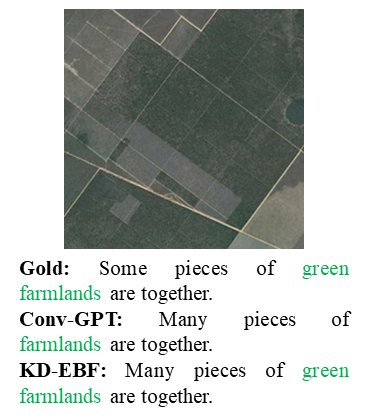}}
  
  \subfloat[\label{example4}]{\includegraphics[width=0.275\textwidth,height=150px]{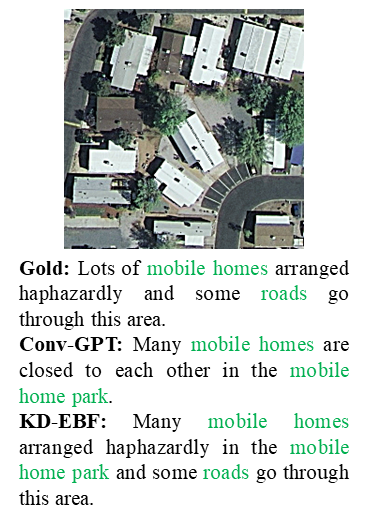}}
  \subfloat[\label{example5}]{\includegraphics[width=0.275\textwidth,height=150px]{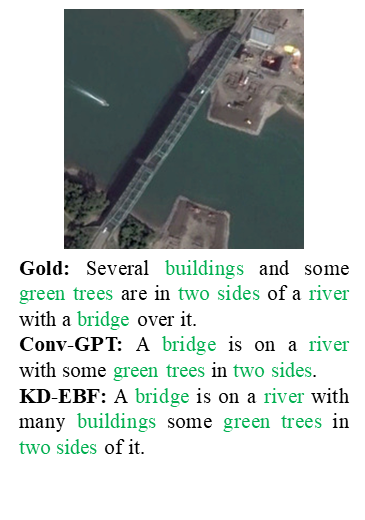}}
  \subfloat[\label{example6}]{\includegraphics[width=0.275\textwidth,height=150px]{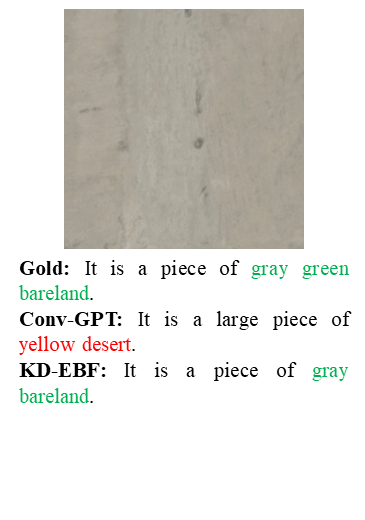}}

  \subfloat[\label{example7}]{\includegraphics[width=0.275\textwidth,height=150px]{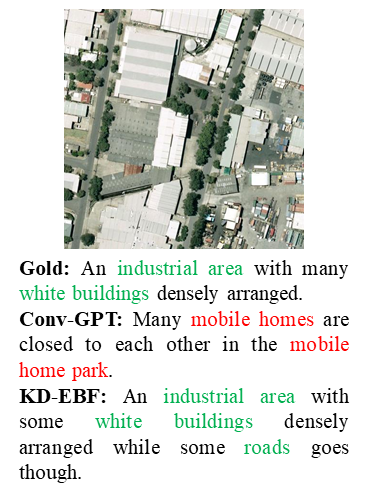}}
  \subfloat[\label{example8}]{\includegraphics[width=0.275\textwidth,height=150px]{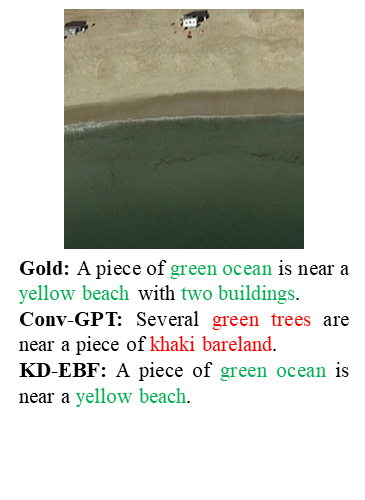}}
  \subfloat[\label{example9}]{\includegraphics[width=0.275\textwidth,height=150px]{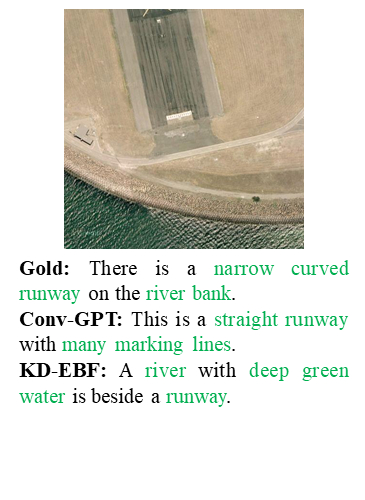}}
   \caption{Examples of RS Image Captioning by Different Methods}
   \label{fig_visual}
\end{figure*}
\subsection{Numerical Evaluation}
\begin{table}[!ht]
\centering
\caption{Comparison of Different RSIC Methods on the SYDNEY dataset}
\label{comp_sydney}
\resizebox{\columnwidth}{!}{
\begin{tabular}{|c|c|c|c|c|c|c|c|}
    \hline
    Edge & BLEU-1 & BLEU-2 & BLEU-3 & BLEU-4 & METEOR & ROUGE-L & CIDEr \\
    \hline
    R-BOW~\cite{lu2017exploring} & 0.5310 & 0.4076 & 0.3319 & 0.2788 & 0.2490 & 0.4922 & 0.7019 \\
    L-FV~\cite{lu2017exploring} & 0.6331 & 0.5332 & 0.4735 & 0.4303 & 0.2967 & 0.5794 & 1.4760 \\
    CSMLF~\cite{wang2019semantic} & 0.4441 & 0.3369 & 0.2815 & 0.2408 & 0.1576 & 0.4018 & 0.9378 \\ 
    CSMLF-FT~\cite{wang2019semantic} & 0.5998 & 0.4583 & 0.3869 & 0.3433 & 0.2475 & 0.5018 & 0.7555 \\
    SD-RSIC~\cite{sumbul2020sd} & 0.7610 & 0.6660 & 0.5860 & 0.5170 & 0.3660 & 0.6570 & 1.6900 \\
    SVM-DBOW~\cite{hoxha2021novel} & 0.7787 & 0.6835 & 0.6023 & 0.5305 & 0.3797 & 0.6992 & 2.2722 \\
    SVM-DCONC~\cite{hoxha2021novel} & 0.7547 & 0.6711 & 0.5970 & 0.5308 & 0.3643 & 0.6746 & 2.2222 \\
    PP~\cite{hoxha2023improving} & 0.7837 & 0.6985 & 0.6322 & 0.5717 & 0.3949 & 0.7106 & 2.5553 \\
    TextGCN~\cite{das2024textgcn} & 0.7680 & 0.6892 & 0.6261 & 0.5786 & 0.4009 & 0.7314 & 2.8595 \\
    TrTr-CMR~\cite{wu2024trtr} & 0.8270 & 0.6994 & 0.6002 & 0.5199 & 0.3803 & 0.7220 & 2.2728 \\    
    Conv-GPT~\cite{das2025good} & 0.7997 & 0.6844 & 0.6325 & 0.5694 & 0.4073 & 0.7349 & 2.4945 \\
    KD-EBF (Proposed) & \textbf{0.8358} & \textbf{0.7505} & \textbf{0.6764} & \textbf{0.6498} & \textbf{0.4579} & \textbf{0.7740} & \textbf{2.8633} \\
    \hline
\end{tabular}}
\end{table}
\begin{table}[!ht]
\centering
\caption{Comparison of Different RSIC Methods on the UCM dataset}
\label{comp_ucm}
\resizebox{\columnwidth}{!}{
\begin{tabular}{|c|c|c|c|c|c|c|c|c|}
    \hline
    Edge & BLEU-1 & BLEU-2 & BLEU-3 & BLEU-4 & METEOR & ROUGE-L & CIDEr\\
    \hline
    R-BOW~\cite{lu2017exploring} & 0.4107 & 0.2249 & 0.1452 & 0.1095 & 0.1098 & 0.3439 & 0.3071 \\
    L-FV~\cite{lu2017exploring} & 0.5897 & 0.4668 & 0.4080 & 0.3683 & 0.2698 & 0.5595 & 1.8438 \\
    CSMLF~\cite{wang2019semantic} & 0.3874 & 0.2145 & 0.1253 & 0.0915 & 0.0954 & 0.3599 & 0.3703 \\
    CSMLF-FT~\cite{wang2019semantic} & 0.3671 & 0.1485 & 0.0763 & 0.0505 & 0.0944 & 0.2986 & 0.1351 \\
    SD-RSIC~\cite{sumbul2020sd} & 0.7480 & 0.6640 & 0.5980 & 0.5380 & 0.3900 & 0.6950 & 2.1320 \\
    SVM-DBOW~\cite{hoxha2021novel} & 0.7635 & 0.6664 & 0.5869 & 0.5195 & 0.3654 & 0.6801 & 2.7142 \\
    SVM-DCONC~\cite{hoxha2021novel} & 0.7653 & 0.6947 & 0.6417 & 0.5942 & 0.3702 & 0.6877 & 2.9228 \\
    PP~\cite{hoxha2023improving} & 0.7973 & 0.7298 & 0.6744 & 0.6262 & 0.4080 & 0.7406 & 3.0964 \\
    TextGCN~\cite{das2024textgcn} & 0.8461 & 0.7844 & 0.7386 & 0.6930 & 0.4868 & 0.8071 & 3.4077 \\
    TrTr-CMR~\cite{wu2024trtr} & 0.8156 & 0.7091 & 0.6220 & 0.5469 & 0.3978 & 0.7442 & 2.4742 \\
    Conv-GPT~\cite{das2025good} & 0.8369 & 0.7712 & 0.7143 & 0.6612 & 0.4566 & 0.8119 & 3.4582 \\
    KD-EBF (Proposed) & \textbf{0.8810} & \textbf{0.8216} & \textbf{0.7675} & \textbf{0.7231} & \textbf{0.5219} & \textbf{0.8502} & \textbf{3.6348} \\
    \hline
\end{tabular}}
\end{table}
\begin{table}[!ht]
\centering
\caption{Comparison of Different RSIC Methods on the RSICD dataset}
\label{comp_rsicd}
\resizebox{\columnwidth}{!}{
\begin{tabular}{|c|c|c|c|c|c|c|c|c|c|}
    \hline
    Edge & BLEU-1 & BLEU-2 & BLEU-3 & BLEU-4 & METEOR & ROUGE-L & CIDEr\\
    \hline
    R-BOW~\cite{lu2017exploring} & 0.4401 & 0.2383 & 0.1514 & 0.1041 & 0.1684 & 0.3605 & 0.4667 \\
    L-FV~\cite{lu2017exploring} & 0.4342 & 0.2453 & 0.1634 & 0.1175 & 0.1712 & 0.3818 & 0.6531 \\
    CSMLF~\cite{wang2019semantic} & 0.5759 & 0.3859 & 0.2832 & 0.2217 & 0.2128 & 0.4455 & 0.5297 \\
    CSMLF-FT~\cite{wang2019semantic} & 0.5106 & 0.2911 & 0.1903 & 0.1352 & 0.1693 & 0.3789 & 0.3388 \\
    SD-RSIC~\cite{sumbul2020sd} & 0.6440 & 0.4740 & 0.3690 & 0.3000 & 0.2490 & 0.5230 & 0.7940 \\
    SVM-DBOW~\cite{hoxha2021novel} & 0.6112 & 0.4277 & 0.3153 & 0.2411 & 0.2303 & 0.4588 & 0.6825 \\
    SVM-DCONC~\cite{hoxha2021novel} & 0.5999 & 0.4347 & 0.3355 & 0.2689 & 0.2299 & 0.4577 & 0.6854 \\
    PP~\cite{hoxha2023improving} & 0.6290 & 0.4599 & 0.3568 & 0.2868 & 0.2530 & 0.4734 & 0.7556 \\
    TextGCN~\cite{das2024textgcn} & 0.6513 & 0.4819 & 0.3747 & 0.3085 & 0.2752 & 0.4804 & 0.8266 \\
    TrTr-CMR~\cite{wu2024trtr} & 0.6201 & 0.3937 & 0.2671 & 0.1932 & 0.2399 & 0.4895 & 0.7518 \\
    Conv-GPT~\cite{das2025good} & 0.6431 & 0.4665 & 0.3602 & 0.3013 & 0.2560 & 0.4945 & 0.8415\\
    KD-EBF (Proposed) & \textbf{0.6781} & \textbf{0.5019} & \textbf{0.3971} & \textbf{0.3242} & \textbf{0.2984} & \textbf{0.5295} & \textbf{0.9196} \\
    \hline
\end{tabular}}
\end{table}
\Cref{comp_sydney,comp_ucm,comp_rsicd} illustrate the comparison between the proposed method (KD-EBF) and several baseline methods in the three datasets. These baseline methods include R-BOW~\cite{lu2017exploring}: RNN with Bag-of-Words feature representations, L-FV~\cite{lu2017exploring}: LSTM with Fisher vector feature representations, CSMLF~\cite{wang2019semantic}: the collective semantic metric learning framework, CSMLF-FT~\cite{wang2019semantic}: the CSMLF with fine-tune, SD-RSIC~\cite{sumbul2020sd}: The summerization-driven RSIC model, SVM-DBOW~\cite{hoxha2021novel}: The SVM-based decoder model with Bag-of-Words feature representations, SVM-DCONC~\cite{hoxha2021novel}: The SVM-based decoder model with concatenating word vector representations, PP~\cite{hoxha2023improving}: the post-processing-based RSIC model, TextGCN~\cite{das2024textgcn}: Text Graph Convolutional Network-based embeddings of words in RSIC model, Conv-GPT~\cite{das2025good}: ConvNext-based encoder and GPT-2-based decoder in the RSIC model, and TrTr-CMR~\cite{wu2024trtr}: Swin transformer-based encoder in the RSIC model. The experimental results indicate the superiority of our method over the other baseline methods.
\subsection{Subjective Evaluation}
\begin{table}[!ht]
\centering
\caption{Subjective Evaluation of Different CNNs on Three Datasets (in \%)}
\label{subjective}
\resizebox{\columnwidth}{!}{%
\begin{tabular}{|c|c|c|c|c|c|c|c|c|c|}
\hline
\multirow{2}{*}{Method} & \multicolumn{3}{c|}{SYDNEY} & \multicolumn{3}{c|}{UCM} & \multicolumn{3}{c|}{RSICD} \\ 
\cline{2-10}
& R & W & U & R & W & U & R & W & U \\
\hline
Conv-GPT~\cite{das2025good} & 89.65 & 3.45 & 6.90 & 92.86 & 3.81 & 3.33 & 83.63 & 6.95 & 9.42 \\
KD-EBF (Proposed) & 91.38 & 3.45 & 5.17 & 94.28 & 1.91 & 3.81 & 84.63 & 7.32 & 8.05 \\
\hline
\end{tabular}%
}
\end{table}
The image captioning task differs significantly from other popular tasks (such as classification), where numerical evaluation is typically sufficient to assess performance. In contrast, image captioning is inherently subjective, as a single image can have multiple valid captions. This subjectivity limits the effectiveness of using a fixed number of reference captions for evaluation. In addition to quantitative analysis, RSIC requires subjective evaluation to better assess a model's performance. This evaluation aims to reflect the human judgment of the generated captions. For this purpose, a human annotator\footnote{The annotator is a highly skilled professional with comprehensive training in working with RSIC models} was employed. The evaluation is based on three predefined labels~\cite{das2025good,lu2017exploring,das2024textgcn}:

\begin{itemize}
    \item \textbf{Related:} The generated caption correctly identifies the main objects and conveys the core meaning of the image with minimal or no errors.
    \item \textbf{Partially Related:} The caption identifies the main objects but fails to convey the image's meaning fully or contains major issues.
    \item \textbf{Unrelated:} The caption is completely irrelevant to the content of the image.
\end{itemize}
\Cref{subjective} compare two models from the eye of a human. The baseline Conv-GPT~\cite{das2025good} is the baseline transformer-based model and KD-EBF is the proposed model. These results clearly show the superiority of our method visually.
\subsection{Visual Examples}
\Cref{fig_visual} presents visual examples of the test data in the three benchmark datasets to compare the performance of the baseline Conv-GPT~\cite{das2025good} model and the proposed KD-EBD model. Each example is accompanied by the corresponding gold (reference) caption to facilitate qualitative assessment. The selected examples illustrate both clear failure cases of the baseline and cases where the proposed model yields more detailed or semantically enriched descriptions.

Several examples highlight instances where the baseline model generates incorrect or misleading descriptions, while the proposed model produces accurate and contextually appropriate captions. In~\Cref{example2}, Conv-GPT misclassifies~\emph{meadow} as~\emph{baseball field}. In~\Cref{example6},~\emph{bareland} is incorrectly predicted as~\emph{desert}. In~\Cref{example7}, the baseline misidentifies~\emph{industrial area} as~\emph{mobile home park}. In~\Cref{example8}, it mislabels~\emph{beach} as~\emph{bareland} and fails to detect~\emph{ocean}, while both models fail to detect~\emph{buildings}. In each of these cases, the proposed KD-EBD model provides accurate labels aligned with the ground truth.

Other examples demonstrate that while both models generate generally correct descriptions, the proposed model includes additional visual or semantic details that improve caption quality. In~\Cref{example1}, KD-EBD identifies~\emph{road}, which is not mentioned in the baseline output. In~\Cref{example3}, it captures the color of~\emph{farmland}. In~\Cref{example4}, the~\emph{road} is again detected only by KD-EBD. In~\Cref{example5}, the proposed model additionally identifies~\emph{buildings}, which are absent from the baseline caption. In~\Cref{example9}, while both models provide largely correct captions, KD-EBD includes~\emph{river}, whereas Conv-GPT focuses on~\emph{marking lines} on~\emph{runway}.
\section{Conclusion}
\label{conclusion}
Although transformer-based models have significantly advanced the field of remote sensing image captioning (RSIC), several key challenges remain unresolved. One critical issue is the substantial complexity of these models, primarily due to the use of separate encoder and decoder transformers. Hence, this structure increases the complexity of the RSIC model and limits real-world applications. Another notable limitation of the existing RSIC methods is the inability to overlook low-level structural cues, which is essential for accurately interpreting scenes with complex layouts or visually similar objects. To overcome these issues, a lightweight transformer-based encoder is developed by reducing the number of parameters in the original encoder architecture. Then, the knowledge distillation technique is applied to transfer the knowledge of the actual model to the proposed lightweight model to improve the performance. In addition, a fusion technique is applied based on the edge detection technique, which combines the original image and the edge-enhanced version. This setup helps the model capture semantic content and structural details, improving its ability to distinguish between similar objects and understand complex spatial arrangements more effectively. Despite having fewer parameters, the experimental results show that the proposed method surpasses several state-of-the-art approaches in RSIC. The lightweight and structurally enhanced design of this model makes it well-suited for efficient and accurate remote sensing image captioning in real-time applications with limited computational resources. Despite these improvements, this work suffers from some shortcomings. Because of the limitation in the training data within the dataset, the model can create inaccurate captions for the classes of images whose class is not present during the model's training. Additionally, the lightweight model may sacrifice some representational capacity, potentially limiting its performance in highly complex scenes. In future work, we aim to address these limitations and improve the model's robustness and generalization.

\bibliographystyle{IEEEtran}
\bibliography{MyWork}

\end{document}